﻿\documentclass[runningheads]{llncs}
\newcommand{\ManuscriptTitle}{MC-TRCM: Observation-Aware Recursive Fusion for Incomplete Mobile and Wearable Mental-Health Feature Views}
\newcommand{\ModelName}{MC-TRCM}
\newcommand{\ModelFullName}{Modality-Conditioned Temporal Recursive Context Model}

\usepackage[T1]{fontenc}
\usepackage{lmodern}
\usepackage{microtype}
\usepackage{graphicx}
\usepackage{amsmath}
\usepackage{amssymb}
\usepackage{booktabs}
\usepackage{array}
\usepackage{siunitx}
\usepackage{tabularx}
\usepackage{threeparttable}
\usepackage{makecell}
\usepackage{ragged2e}
\usepackage{xurl}
\usepackage{color}

\newcolumntype{L}[1]{>{\RaggedRight\arraybackslash}p{#1}}
\newcolumntype{C}[1]{>{\Centering\arraybackslash}p{#1}}
\newcolumntype{Y}{>{\RaggedRight\arraybackslash}X}
\newcolumntype{Z}{>{\Centering\arraybackslash}X}
\newcommand{\best}[1]{\textbf{#1}}
\newcommand{\msd}[2]{\makecell[c]{#1\\[-0.4ex]{\tiny $\pm$ #2}}}

\allowdisplaybreaks[1]
\begin{document}

\title{\ManuscriptTitle}
\titlerunning{MC-TRCM for Mental-Health Feature Views}

\author{Wentao Wang\inst{1}\fnmsep\thanks{Wentao Wang and Lifeng Han contributed equally to this work and share first authorship.} \and
Lifeng Han\inst{1,\star} \and
Zining Ren\inst{2} \and
Hengyu Zhong\inst{3} \and
Guangyu~Zou\inst{1}\fnmsep\thanks{Corresponding author.}}
\authorrunning{W. Wang et al.}

\institute{Dalian University of Technology, Dalian, China\\
\email{\{shiyanxi1,2369787465\}@mail.dlut.edu.cn, gyzou@dlut.edu.cn} \and
The Hong Kong Polytechnic University, Hong Kong, China\\
\email{25094416g@connect.polyu.hk} \and
Southwest University, Chongqing, China\\
\email{hengyuzhong@email.swu.edu.cn}}

\maketitle

\begin{abstract}
Public mobile and wearable mental-health datasets often provide summarized feature tables rather than synchronized raw sensor streams. In these releases, each anchor corresponds to a survey or label time and may combine phone or wearable summaries, prior symptom scores, demographics, clinical variables, and source-availability indicators. We propose the \ModelFullName{} (\ModelName{}), which preserves each feature source as a separate token and incorporates missingness as part of the input context. Observed sources are encoded with values and missingness summaries, absent sources use learned absence tokens, dataset and task embeddings condition fusion, and a recursive prediction head refines each output over validation-selected steps. We evaluated \ModelName{} on six predefined endpoints from DepreST-CAT and Prediction of Severity Change-Depression (PSYCHE-D) using participant-level splits and validation-only model selection. \ModelName{} achieved the lowest mean absolute error on DepreST-CAT Patient Health Questionnaire-9 (PHQ-9) and Generalized Anxiety Disorder-7 (GAD-7) severity, improving over the best tabular reference by 0.181 and 0.217 scale points. Classification endpoints showed task-dependent behavior: \ModelName{} matched the best rounded GAD-7 category balanced accuracy, was numerically highest by 0.002 balanced-accuracy points on PSYCHE-D multiclass prediction, and remained close to the strongest references on PHQ-9 category and PSYCHE-D binary prediction. Ablations support Feature-wise Linear Modulation, absence tokens, missingness projections, and recursive refinement, while calibration and feature-source controls characterize endpoint behavior. Our code is available at \url{https://github.com/Botwwt/MC-TRCM}.
\end{abstract}

\section{Introduction}
Depression assessment continues to depend on episodic self-report and clinical encounters. Symptom fluctuation, recall bias, gaps in clinical contact, and changing adherence can leave clinically important intervals partially observed. Mobile and wearable sensing can add behavioral context between assessments \cite{deangel2022digital,leaning2024smartphone,abdalrazaq2023wearable}, but these data do not directly measure depression. They provide indirect proxies for sleep, activity, mobility, communication, and device interaction, and the absence of a proxy can itself be informative.

In reproducible public studies, the modeling problem is increasingly a feature-view rather than raw-stream setting. Public datasets such as StudentLife, DepreST-CAT, Prediction of Severity Change-Depression (PSYCHE-D), Depresjon, and OBF-Psychiatric usually release processed views rather than raw, synchronized sensor streams \cite{wang2014studentlife,tlachac2022deprest,makhmutova2022predicting,berle2010actigraphic,garciaceja2025obf}. These views mix communication summaries, mobility statistics, sleep and activity aggregates, symptom history, demographic variables, clinical context, and availability indicators. A missing sleep window, communication record, or clinical field may reflect non-wear, study disengagement, hospitalization, privacy settings, or release constraints. The core modeling challenge is to learn jointly from recorded values, source identity, and the pattern of available and missing sources.

Existing solutions span regularized regression, gradient-boosted trees, Explainable Boosting Machine (EBM) models, multilayer perceptron (MLP) models, recurrent networks, and Transformer encoders \cite{friedman2010glmnet,chen2016xgboost,ke2017lightgbm,nori2019interpretml,vaswani2017transformer}. These models are strong predictors, but flat concatenation treats mobility, communication, baseline symptoms, and missingness indicators as exchangeable columns. Conventional imputation, as commonly used in benchmark pipelines, replaces absent entries before modeling and thereby treats missingness primarily as a preprocessing artifact \cite{donders2006gentle,van2011mice}. A single fused representation may not transfer uniformly across labels: Patient Health Questionnaire-9 (PHQ-9) severity, Generalized Anxiety Disorder-7 (GAD-7) severity, and PHQ-change labels differ in scale and in the clinical question they encode. This distinction matters because depression- and anxiety-related symptoms are heterogeneous latent constructs, and released sensing features are indirect behavioral proxies rather than direct clinical measurements.

To address this setting, we propose the \ModelFullName{} (\ModelName{}). \ModelName{} treats each feature source as a token and conditions fusion on dataset and task identity. This design lets the encoder use both measured values and availability patterns without collapsing all sources into a single flat vector. Missingness and source availability enter through missingness-aware gates, learned absence tokens, and missingness projections. Feature-wise Linear Modulation (FiLM) uses dataset and task embeddings to adapt the representation to cohort-specific feature distributions and endpoint identity, such as DepreST-CAT communication summaries and PSYCHE-D wearable-derived change labels. Recursive decoding then refines each endpoint prediction over a shared task-conditioned representation of the anchor.

We evaluate \ModelName{} on six DepreST-CAT and PSYCHE-D severity and classification endpoints using participant-level splits. All model selection, early stopping, calibration, thresholding, class-bias correction, recursive-depth selection, and hyperparameter selection use training and validation data only. \ModelName{} achieves the lowest mean absolute error (MAE) on both DepreST-CAT severity endpoints among the evaluated models. Classification results are task-dependent, with rounded ties or small numerical differences from the strongest references. \textbf{Our contributions are as follows.}
\begin{itemize}
\item \textbf{An observation-aware formulation and recursive architecture.} We formulate incomplete public mental-health releases as feature-view learning problems where observed values, source identities, and availability patterns are modeled jointly through source tokens, missingness projections, dataset- and task-conditioned FiLM, and recursive decoding.

\item \textbf{A leakage-controlled benchmark with explicit source accounting.} We evaluate six predefined DepreST-CAT and PSYCHE-D tasks, covering 360 DepreST-CAT participants and 10,866 PSYCHE-D task rows, using participant-level splits, training-only preprocessing, and validation-only choices for model selection, calibration, thresholding, class-bias correction, recursive depth, and hyperparameters. The protocol also reports matched feature-source controls, separating sensor and missingness views from symptom, static, and clinical context.

\item \textbf{Numerical improvements and component evidence.} \ModelName{} reduces MAE by 0.181 PHQ-9 points and 0.217 GAD-7 points relative to the strongest tabular references, matches the best rounded GAD-7 category BA, and is numerically highest on PSYCHE-D multiclass by 0.002 BA. Five-seed ablations support the contributions of missingness modeling, FiLM conditioning, and recursive refinement.
\end{itemize}

\section{Related Work}
\subsection{Digital Phenotyping Benchmarks, Metrics, and Calibration}
Mobile and wearable depression-related studies complement questionnaires with activity, sleep, mobility, phone-use, and communication signals, but performance depends on cohort composition, label timing, feature construction, and validation design \cite{jacobson2019digital,deangel2022digital,leaning2024smartphone,abdalrazaq2023wearable}. The public benchmarks discussed above enable reproducible comparison through fixed or reconstructable feature views \cite{wang2014studentlife,tlachac2022deprest,makhmutova2022predicting,berle2010actigraphic,garciaceja2025obf}.

Metric choice should match the endpoint being modeled. Questionnaire severity requires an error measure that remains interpretable in scale points, whereas imbalanced category and change labels require class-balanced evaluation rather than raw accuracy. Calibration is also relevant because a correct class decision can still be assigned with unreliable confidence \cite{brier1950verification,guo2017calibration}. These considerations motivate the MAE, balanced accuracy (BA), Brier score, and expected calibration error (ECE) protocol defined below, together with validation-only thresholding and calibration.

\subsection{Generalization, Missingness, and Feature Views}
Cross-study mobile sensing work shows that generalization can shift across cohorts, devices, years, and sensing protocols \cite{adler2022passive,xu2022globem}. This is important for mental-health benchmarks, where the observed data are indirect proxies and missingness may reflect both measurement and behavior. Zero and mean filling, KNN and SVD imputation, and MICE complete released matrices before modeling, which can obscure the distinction between an observed low value, a partially reliable source, and an unavailable source \cite{donders2006gentle,troyanskaya2001missing,van2011mice}. MC-TRCM keeps feature sources separate, represents availability patterns explicitly, and evaluates matched feature-source controls.

\subsection{Tabular, Neural, and Conditioned Fusion Models}
Prepared releases are often treated as tabular matrices, making regularized regression, XGBoost, LightGBM, and EBM strong references \cite{friedman2010glmnet,chen2016xgboost,ke2017lightgbm,nori2019interpretml}. Neural sequence and set encoders such as gated recurrent unit (GRU), long short-term memory (LSTM), and Transformer models provide flexible fusion \cite{hochreiter1997lstm,cho2014gru,vaswani2017transformer}. Public depression-related feature-view releases add a complementary requirement: modality, endpoint, and availability-pattern structure should be preserved even when raw streams are not released. FiLM conditioning shifts and scales representations with context \cite{perez2018film}, while PCGrad addresses task gradient conflict in multitask learning \cite{yu2020pcgrad}. MC-TRCM combines these ideas by representing each source as a token, retaining modality absence as context, adapting fusion with dataset and task conditioning, and updating endpoint-specific task states recursively.

\section{Methods}
\begin{figure}[!t]
\centering
\includegraphics[width=\linewidth,height=0.25\textheight,keepaspectratio]{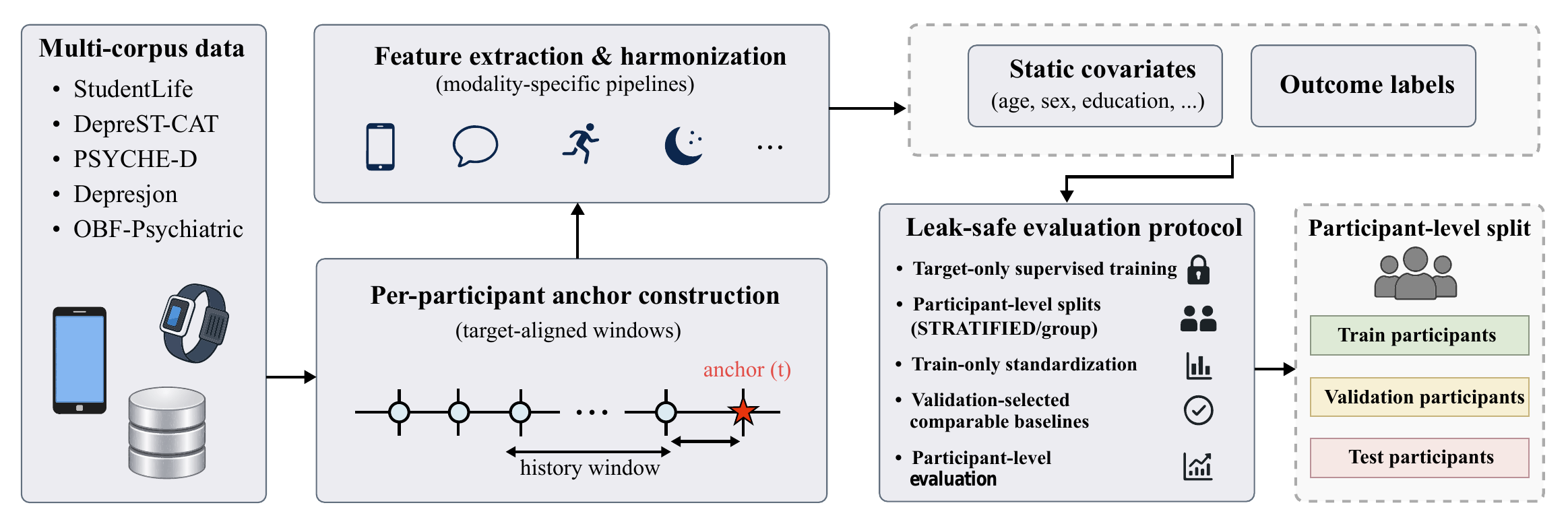}
\caption{Endpoint construction and evaluation protocol. Public cohorts are harmonized into modality-specific feature views, aligned to participant anchors, and evaluated with training-only preprocessing and validation-locked model selection.}
\label{fig:mctrcm_protocol}
\end{figure}

We formulate participant-level, endpoint-specific prediction from heterogeneous feature views and introduce \ModelName{}. The model addresses separate feature groups, informative missingness, and labels spanning symptom scores, categories, and changes. Figure~\ref{fig:mctrcm_protocol} summarizes data construction and evaluation.
\subsection{Problem Formulation}
Let $d$ denote a dataset, $i$ a participant, $a$ an anchor, $\tau$ an endpoint, and $m\in\mathcal{M}_d$ a modality or feature source in the fixed released source set for dataset $d$. An anchor is a survey time, label time, or released processed row paired with features available at or before that anchor. For anchor $a$, endpoint $\tau$ has label $y_{a,\tau}$ when available. Repeated anchors from the same participant are assigned to exactly one train, validation, or test split, so no participant contributes rows to multiple splits.

The model observes standardized modality values $\tilde{x}_{a,m}$, missingness summaries $q_{a,m}$, and an observed indicator $s_{a,m}\in\{0,1\}$. Here, $\tilde{x}_{a,m}$ represents the released behavioral or contextual measurements for an observed feature source after training-split imputation and standardization, $q_{a,m}$ summarizes column-level masks, missing ratios, recency proxies, or source-availability indicators, and $s_{a,m}$ records whether the source is available at all for the anchor. The learning problem is multitask prediction over regression, binary, ordinal, and multiclass endpoints while respecting endpoint-specific label availability.
\subsection{Feature Views and Endpoints}
We organize feature columns into five sources: sensor values, missingness, symptom context, static demographics, and clinical context. Sensor values include activity, sleep, communication, phone use, and mobility summaries when available. Missingness features include observed indicators, missing ratios, recency proxies, and modality masks. Symptom context includes temporally prior variables such as baseline PHQ-9 in PSYCHE-D or pre-study PHQ-9 in StudentLife. Static demographics include released age and sex variables. Clinical context includes prior treatment or baseline clinical variables when they are available before the modeled label.
The primary benchmark contains six predefined endpoints: DepreST-CAT PHQ-9 severity, DepreST-CAT GAD-7 severity, DepreST-CAT PHQ-9 category, DepreST-CAT GAD-7 category, PSYCHE-D PHQ-change binary, and PSYCHE-D PHQ-change multiclass. DepreST-CAT uses 216, 72, and 72 participants for training, validation, and testing with 283 released features; PSYCHE-D uses 2405, 808, and 823 participants, plus 6494, 2119, and 2253 task rows across the same splits, with 325 features. StudentLife, Depresjon, and OBF-Psychiatric are auxiliary checks: StudentLife has 7 test participants, Depresjon has 11 status and 4 Montgomery-Asberg Depression Rating Scale (MADRS) test participants, and OBF-Psychiatric has 33 held-out participants for clinical-vs-control classification \cite{garciaceja2025obf}. These auxiliary endpoints provide external checks rather than the primary basis for model ranking. For example, OBF-Psychiatric reaches BA = 1.000 for both \ModelName{} and LightGBM, so it mainly confirms that the task is saturated under this split. The locked primary metrics are MAE for continuous severity and BA for classification and ordinal endpoints; Brier score and ECE are used for calibration analyses. For ordinal endpoints, BA is class-balanced discrimination rather than ordered-error evaluation. Endpoint choices follow the original cohort framing \cite{wang2014studentlife,tlachac2022deprest,makhmutova2022predicting,berle2010actigraphic,garciaceja2025obf}.
\subsection{\ModelName{} Architecture}
\begin{figure}[!t]
\centering
\includegraphics[width=\linewidth,height=0.27\textheight,keepaspectratio]{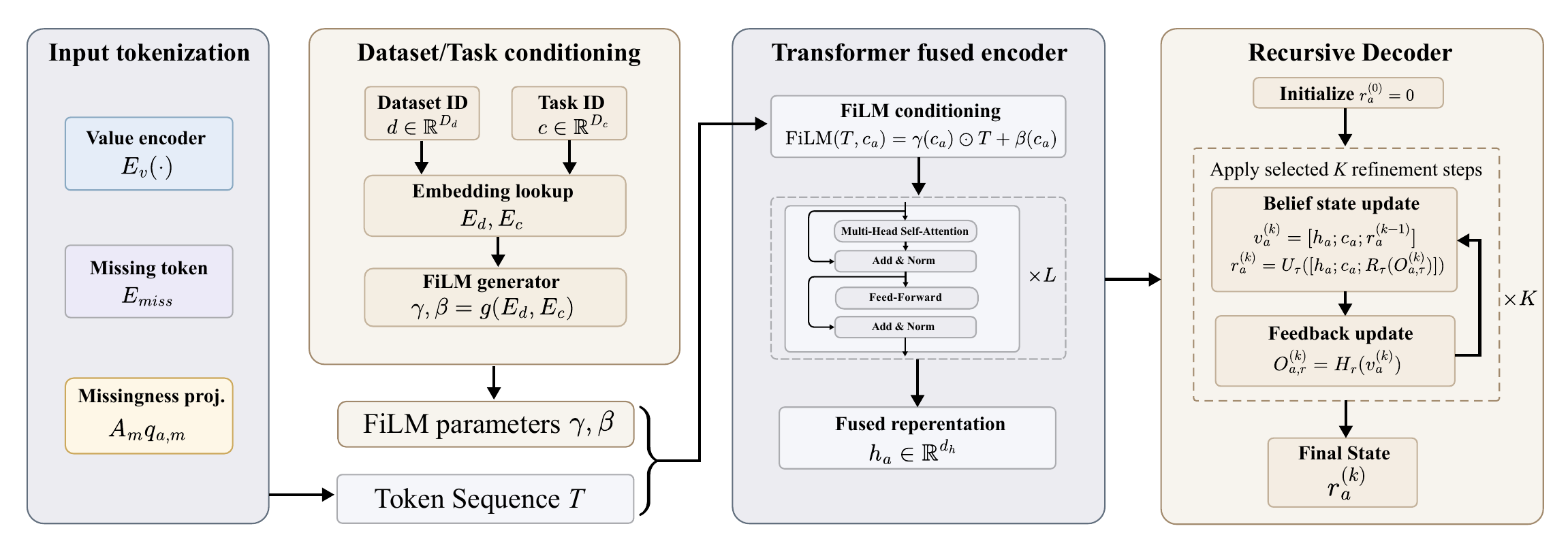}
\caption{MC-TRCM architecture. Source tokens carry observed values, missingness context, and modality identity; FiLM-conditioned Transformer fusion and recursive decoding produce the endpoint-specific prediction.}
\label{fig:mctrcm_architecture}
\end{figure}

The architecture in Figure~\ref{fig:mctrcm_architecture} takes these anchored feature views as missingness-aware source tokens, conditions fusion on dataset and endpoint identity, and refines each prediction recursively.

\paragraph{Modality tokenization and missingness-aware gating.}
Public depression-related datasets release modular views whose measurement processes differ across communication, mobility, sleep, prior symptoms, demographics, and clinical context. MC-TRCM encodes each feature source as a tokenized view, so the fusion layer receives source objects rather than an unordered collection of columns.

For modality $m$, an encoder $E_m$ maps standardized observed features into a shared token space, and a gate $G_m$ controls how much of that observed-value representation should enter the token. The value encoder and gate are used when the source is observed:
\begin{equation}
t_{a,m} =
\begin{cases}
\sigma(G_m([\tilde{x}_{a,m};q_{a,m}]))\odot E_m(\tilde{x}_{a,m}) + A_m q_{a,m}+e_m, & s_{a,m}=1,\\
u_m + A_m q_{a,m} + e_m, & s_{a,m}=0,
\end{cases}
\label{eq:missing-token}
\end{equation}
where $A_m q_{a,m}$ is a learned missingness projection, $e_m$ is a source embedding, and $u_m$ is a learned absence token. When $s_{a,m}=0$, the value encoder and gate are not invoked; the token is formed from the learned absence token, the missingness summary, and the source identity. For partially missing observed sources, $\tilde{x}_{a,m}$ is the vector after training-only imputation and standardization, while column-level masks, missing ratios, and recency proxies enter through $q_{a,m}$.

The key tensor dimensions are
\begin{equation}
\begin{gathered}
\tilde{x}_{a,m}\in\mathbb{R}^{p_m},\quad
q_{a,m}\in\mathbb{R}^{r_m},\quad
t_{a,m},u_m,e_m\in\mathbb{R}^{d_h},\quad
r_{a,\tau}^{(k)}\in\mathbb{R}^{d_r},\\
E_m:\mathbb{R}^{p_m}\to\mathbb{R}^{d_h},\quad
G_m:\mathbb{R}^{p_m+r_m}\to\mathbb{R},\quad
A_m\in\mathbb{R}^{d_h\times r_m}.
\end{gathered}
\end{equation}
In the implementation, $G_m$ outputs a scalar gate that is broadcast over the token dimension. This separation distinguishes observed values, reliability or availability patterns, and source identity without treating missingness as an imputation artifact.

\paragraph{FiLM dataset and task conditioning.}
Cross-dataset and cross-task heterogeneity is a central design constraint. DepreST-CAT communication summaries, PSYCHE-D processed wearable features, and actigraphy-based cohorts occupy different feature distributions and are paired with different labels. MC-TRCM uses dataset and task embeddings to make this heterogeneity explicit.

Dataset and task embeddings form a context vector
\begin{equation}
c_{a,\tau}=[e^d_{d(a)};e^\tau_\tau].
\end{equation}
The reported configuration applies FiLM both before token fusion and after Transformer fusion, as selected on validation data and fixed before test evaluation:
\begin{equation}
\mathrm{FiLM}_{\ell}(T,c_{a,\tau})=\gamma_{\ell}(c_{a,\tau})\odot T+\beta_{\ell}(c_{a,\tau}).
\label{eq:film}
\end{equation}
The source tokenization $T_a=\{t_{a,m}:m\in\mathcal{M}_d\}$ is shared, but FiLM-conditioned fusion and decoding are run separately for each observed anchor--task pair. The final model uses two Transformer blocks with hidden dimension $d_h=64$, four attention heads, feed-forward dimension 160, dropout 0.1, and a projection to a 96-dimensional latent representation. Mean pooling includes observed-source tokens and learned absence tokens, excluding padding tokens introduced for cross-dataset alignment; thus the fused representation can use the availability pattern as input context. With $T_{a,\tau}^{(2)}$ denoting the task-conditioned token sequence after the second block,
\begin{equation}
h_{a,\tau}=f_\theta(T_a,c_{a,\tau})
= W_h\left(\frac{1}{|\mathcal{M}_d|}\sum_{m\in\mathcal{M}_d}T_{a,\tau,m}^{(2)}\right).
\label{eq:fusion}
\end{equation}
The rationale of FiLM is dynamic conditioning: $\gamma_{\ell}(c_{a,\tau})$ scales token dimensions and $\beta_{\ell}(c_{a,\tau})$ shifts them, allowing the same architecture to adjust its internal representation when the dataset or endpoint changes. Statistically, $e^d_{d(a)}$ encodes the cohort and release protocol, while $e^\tau_\tau$ encodes the clinical question being asked. The resulting representation $h_{a,\tau}$ summarizes observed values, source identity, and missingness context conditioned on both dataset and task.

\paragraph{Recursive decoding.}
The decoder produces the task output over several refinement steps. At each step, it combines the task-conditioned fused representation, the task identity, and feedback from the previous prediction to update an internal decoder state. The final step gives the reported prediction. This differs from simply increasing encoder depth because recurrence is applied at the task head, passes the current endpoint estimate through $R_\tau$, and supervises intermediate states through the training objective.

For dataset $d$, recursive depth $K_d$ is selected on validation data. The recursive decoder initializes $r_{a,\tau}^{(0)}=\mathbf{0}$. For $k=1,\ldots,K_d$,
\begin{align}
v_{a,\tau}^{(k)} &= [h_{a,\tau};c_{a,\tau};r_{a,\tau}^{(k-1)}],\\
o_{a,\tau}^{(k)} &= H_\tau(v_{a,\tau}^{(k)}),\\
r_{a,\tau}^{(k)} &= U_\tau\!\left([h_{a,\tau};c_{a,\tau};R_\tau(o_{a,\tau}^{(k)})]\right),
\label{eq:recursive-update}
\end{align}
where $H_\tau$ is a task head, $R_\tau(o)=B_\tau\phi_\tau(o)$ maps task outputs into a feedback space, and $U_\tau$ updates the task-specific recursive state. We use $\phi_\tau(o)$ as the scalar prediction for regression, the probability or logit vector for binary endpoints, and the class-probability vector for ordinal or multiclass endpoints. The recursive state is an internal decoder variable used for prediction refinement rather than a clinically validated latent state. The selected refinement depth defines the final prediction, $\hat y_{a,\tau}=o_{a,\tau}^{(K_d)}$. $K_d$ is selected from $\{1,2,4,6,8\}$; $K_d=1$ corresponds to a single-step decoder, while larger values allow iterative refinement when the validation evidence supports it.

For depth selection, each candidate $K$ is scored on validation data within dataset $d$ as $S_d(K)=|\mathcal{T}_d|^{-1}\sum_{\tau\in\mathcal{T}_d}s_{d,\tau}(K)$, where $s_{d,\tau}(K)=1-\mathrm{MAE}_{d,\tau}(K)/R_\tau$ for severity endpoints and $s_{d,\tau}(K)=\mathrm{BA}_{d,\tau}(K)$ for classification or ordinal endpoints. Here $R_\tau$ is the questionnaire range, 27 for PHQ-9 and 21 for GAD-7. The normalized severity score is used only for validation selection; final severity results are reported as raw MAE in questionnaire scale points.
\subsection{Training Objective and Optimization}
Within each dataset run, training accumulates losses over recursive steps using normalized increasing weights $w_k=k/\sum_{\ell=1}^{K_d}\ell$:
\begin{equation}
\mathcal{L}_{\mathrm{task}} =
\frac{1}{|\mathcal{T}_B|}
\sum_{\tau\in\mathcal{T}_B}
\left[
\exp(-\lambda_\tau)\sum_{k=1}^{K_d} w_k L_\tau^{(k)}
+\lambda_\tau
\right],
\end{equation}
where $\mathcal{T}_B$ is the set of tasks observed in minibatch $B$, $\lambda_\tau$ is a learned task uncertainty weight, and $L_\tau^{(k)}$ is the endpoint-specific loss at recursive step $k$. The increasing recursive weights make later refinements more influential while still supervising intermediate states, which stabilizes training when early recursive predictions are coarse. Regression tasks compare Huber, MSE, and hybrid MAE-Huber losses in validation search. Binary and multiclass tasks include class-balanced cross entropy, focal loss, and label-smoothing options when selected by validation. Ordinal tasks use ordinal decoding with validation-fitted class-bias adjustment for category prediction.

The shared encoder is trained across related but nonidentical clinical endpoints, producing gradient geometry in which PHQ-9 severity, GAD-7 severity, category prediction, and PHQ-change classification may emphasize different symptom dimensions. The selected full configuration includes PCGrad as the validation-selected conflict-handling option. PCGrad projects a task gradient away from another task's gradient when their inner product is negative \cite{yu2020pcgrad}. Optimizer variants were explored during validation but are not used as primary architectural evidence. We train with AdamW, gradient clipping, modality dropout candidates, and early stopping on validation primary metrics.
\subsection{Baselines, Calibration, and Metrics}
The benchmark compares \ModelName{} with Elastic Net or logistic regression, LightGBM, XGBoost, EBM, MLP, GRU, LSTM, Transformer, and a shared encoder multitask MLP; null predictors are retained as sanity checks. EBM serves as a strong, high-capacity glass-box reference for clinically inspectable additive and pairwise tabular structure. Baselines use the same participant-level splits, endpoint definitions, feature matrices, training-only preprocessing, and validation-only selection protocol. Elastic Net, logistic regression, and MLP baselines use median imputation followed by standardization fitted on the training split before model fitting. Tree-based and EBM references use validation-selected missing-value handling; whenever imputation is required, imputation parameters are fitted on the training split and applied unchanged to validation and test splits, with missingness indicators retained when available. Neural references are capacity-matched where possible, using the same input feature matrices and validation metric, so performance differences are not primarily attributable to unmatched model capacity. The GRU, LSTM, and Transformer baselines use a fixed four-token sequence ordered as short, medium, long, and static feature views from the same released matrices; within each token, base features follow a fixed sorted column order, and these baselines remove dataset/task FiLM and recursive decoding. For each endpoint, tabular and neural reference families are selected by validation primary metric before test metrics are summarized.

Continuous severity endpoints report MAE, the mean absolute difference between predicted and observed PHQ-9, GAD-7, or MADRS scores; one MAE unit corresponds to one questionnaire scale point, and lower is better. Classification and ordinal endpoints report BA, the average of class-wise recalls; each class contributes equally under imbalance, and higher is better. Calibration analyses use Brier score and ECE. For binary endpoints, Brier score is $\frac{1}{N}\sum_{i=1}^{N}(p_i-y_i)^2$ using the positive-class probability. For multiclass and ordinal endpoints, Brier score and ECE are
\begin{equation}
\begin{gathered}
\mathrm{Brier}=\frac{1}{N}\sum_{i=1}^{N}\sum_{c=1}^{C}(p_{i,c}-\mathbf{1}\{y_i=c\})^2,\\
\mathrm{ECE}=\sum_{b=1}^{B}\frac{|I_b|}{N}\left|\mathrm{acc}(I_b)-\mathrm{conf}(I_b)\right|.
\end{gathered}
\end{equation}
ECE uses $B=10$ equal-width confidence bins, with confidence defined as $\max_c p_{i,c}$.
Binary thresholds, ordinal and multiclass class-bias corrections, and temperature scaling are fitted on validation data, excluding test labels. All models are evaluated over five seeds when applicable; deterministic fits with fixed post-processing report zero standard deviation. Continuous predictions and class probabilities are evaluated for each \ModelName{} seed before applying validation-fitted threshold or class-bias calibration.

\subsection{Ablations and Robustness}
Ablations remove one component at a time from the selected full model and are run for five seeds: no FiLM conditioning, no absence token, no missingness projection, no PCGrad, no task conditioning, and no recursive refinement. The no-absence-token variant removes $u_m$, whereas the no-missingness-projection variant removes $A_m q_{a,m}$. Optimizer variants are validation sensitivity checks rather than primary architectural ablations. Recursive-depth sensitivity, calibration comparisons, and matched feature-source controls interpret robustness across sensor values, missingness, symptom context, static demographics, clinical context, and combined views.
\section{Experiments and Results}
This section reports the six predefined endpoint metrics, feature-source accounting, calibration, recursive depth, and ablations. DepreST-CAT and PSYCHE-D define the primary benchmark because they provide the largest public endpoint set in this study. StudentLife, Depresjon, and OBF-Psychiatric are run in the same experimental suite as auxiliary checks. The results provide architecture-level evidence under fair alignment while preserving transparent feature-source accounting.
\subsection{Main Benchmark Results}

\begin{table}[!t]
\centering
\caption{Main test performance across pre-specified comparator families (mean and standard deviation).}
\label{tab:core_main_comparison}
\scriptsize
\setlength{\tabcolsep}{1.5pt} 
\renewcommand{\arraystretch}{1.05}
\begin{tabularx}{\linewidth}{@{}L{0.18\linewidth}*{6}{Z}@{}} 
\toprule
Model & \makecell[c]{PHQ-9\\MAE$\downarrow$} & \makecell[c]{GAD-7\\MAE$\downarrow$} & \makecell[c]{PHQ-9\\BA$\uparrow$} & \makecell[c]{GAD-7\\BA$\uparrow$} & \makecell[c]{Psy. bin.\\BA$\uparrow$} & \makecell[c]{Psy. multi.\\BA$\uparrow$} \\
\midrule
Elastic Net & \msd{7.582}{0.000} & \msd{6.791}{0.000} & \msd{0.178}{0.000} & \msd{0.245}{0.000} & \msd{0.582}{0.001} & \msd{0.499}{0.000} \\
LightGBM & \msd{6.328}{0.090} & \msd{5.297}{0.084} & \msd{0.239}{0.043} & \best{\msd{0.343}{0.031}} & \msd{0.574}{0.004} & \msd{0.488}{0.002} \\
XGBoost & \msd{5.918}{0.066} & \msd{5.318}{0.096} & \msd{0.242}{0.024} & \msd{0.297}{0.019} & \msd{0.587}{0.005} & \msd{0.505}{0.005} \\
EBM & \msd{5.817}{0.085} & \msd{5.028}{0.088} & \msd{0.220}{0.050} & \msd{0.308}{0.030} & \best{\msd{0.589}{0.007}} & \msd{0.512}{0.003} \\
MLP & \msd{6.138}{0.100} & \msd{5.449}{0.106} & \msd{0.193}{0.041} & \msd{0.278}{0.035} & \msd{0.566}{0.003} & \msd{0.463}{0.015} \\
GRU & \msd{6.118}{0.031} & \msd{5.364}{0.003} & \msd{0.212}{0.040} & \msd{0.250}{0.016} & \msd{0.582}{0.004} & \msd{0.509}{0.004} \\
LSTM & \msd{6.229}{0.123} & \msd{5.366}{0.009} & \best{\msd{0.252}{0.031}} & \msd{0.223}{0.027} & \msd{0.575}{0.005} & \msd{0.507}{0.004} \\
Transformer & \msd{6.233}{0.172} & \msd{5.415}{0.167} & \msd{0.240}{0.032} & \msd{0.223}{0.026} & \msd{0.578}{0.011} & \msd{0.497}{0.015} \\
Multitask MLP & \msd{6.131}{0.107} & \msd{5.354}{0.040} & \msd{0.205}{0.046} & \msd{0.272}{0.017} & \msd{0.582}{0.007} & \msd{0.500}{0.007} \\
\ModelName{} & \best{\msd{5.636}{0.035}} & \best{\msd{4.811}{0.044}} & \msd{0.246}{0.041} & \best{\msd{0.343}{0.031}} & \msd{0.582}{0.011} & \best{\msd{0.514}{0.008}} \\
\bottomrule
\end{tabularx}
\par\smallskip
\begin{minipage}{0.96\linewidth}
\footnotesize\textit{Note.} Second line is SD over five seeds when stochasticity or validation-selected post-processing varies; deterministic fits with fixed post-processing report 0.000. Bold marks the best rounded value.
\end{minipage}
\end{table}

\normalsize
\vspace{0.3em}

\begin{table}[!t]
\normalsize
\caption{Auxiliary dataset test results. These endpoints are outside the primary benchmark because held-out sets are small or saturated.}
\label{tab:small_dataset_results}
\scriptsize
\setlength{\tabcolsep}{2pt} 
\renewcommand{\arraystretch}{1.05}
\begin{tabularx}{\linewidth}{@{}L{0.12\linewidth}L{0.17\linewidth}C{0.085\linewidth}C{0.07\linewidth}C{0.15\linewidth}L{0.16\linewidth}Z@{}}
\toprule
  Dataset & Endpoint & Metric & Test $n$ & \ModelName{} & Best ref. & Ref. value \\
\midrule
StudentLife & PHQ-9 severity & MAE$\downarrow$ & 7 & \best{\msd{3.812}{0.302}} & Transformer & \msd{4.916}{0.229} \\
  StudentLife & PHQ-9 category & BA$\uparrow$ & 7 & \best{\msd{0.350}{0.152}} & XGBoost & \msd{0.250}{0.000} \\
  Depresjon & Depr. status & BA$\uparrow$ & 11 & \msd{0.911}{0.124} & EBM & \best{\msd{1.000}{0.000}} \\
  Depresjon & MADRS severity & MAE$\downarrow$ & 4 & \best{\msd{2.001}{0.622}} & EBM & \msd{2.493}{0.198} \\
  OBF & Clin. vs. Ctrl. & BA$\uparrow$ & 33 & \best{\msd{1.000}{0.000}} & LightGBM & \best{\msd{1.000}{0.000}} \\
\bottomrule
\end{tabularx}
\par\smallskip
\begin{minipage}{0.96\linewidth}
  \footnotesize\textit{Note.} Best ref. is the strongest validation-selected comparator. Clin. vs. Ctrl. denotes Clinical vs. control classification. Second line is SD over five seeds; bold marks the best rounded value.
\end{minipage}
\end{table}
Table~\ref{tab:core_main_comparison} reports the pre-specified comparator families on the six core endpoint metrics after validation-selected hyperparameters and recursive depth. Baseline families, thresholding, class-bias correction, and calibration are frozen from training and validation data before test evaluation. The main result is that \ModelName{} is most clearly beneficial for continuous severity estimation. It reduces MAE by 0.181 PHQ-9 points and 0.217 GAD-7 points relative to the strongest tabular references, indicating that observation-aware source modeling is particularly useful when the target is symptom magnitude rather than a discrete category.

Classification results are more endpoint-dependent. \ModelName{} matches the best rounded GAD-7 category BA and is numerically highest on PSYCHE-D multiclass by 0.002 BA, but remains close to the strongest references on PHQ-9 category and PSYCHE-D binary prediction. Thus, the benchmark evidence supports \ModelName{} mainly as a stronger severity-estimation model, while categorical discrimination still depends on endpoint structure, class balance, and threshold behavior.

Table~\ref{tab:small_dataset_results} reports auxiliary endpoints used as external checks rather than the primary basis for ranking. These results follow the same pattern: the largest gains occur on continuous severity tasks, including StudentLife PHQ-9 severity (MAE $3.812\pm0.302$ vs. $4.916\pm0.229$) and Depresjon MADRS severity (MAE $2.001\pm0.622$ vs. $2.493\pm0.198$). In contrast, OBF-Psychiatric reaches BA $=1.000$ for both \ModelName{} and LightGBM, suggesting that this split is saturated and provides little discrimination among strong models.

\subsection{Calibration and Recursive Depth}

\begin{figure}[!t]
\centering
\includegraphics[width=\linewidth,height=0.27\textheight,keepaspectratio]{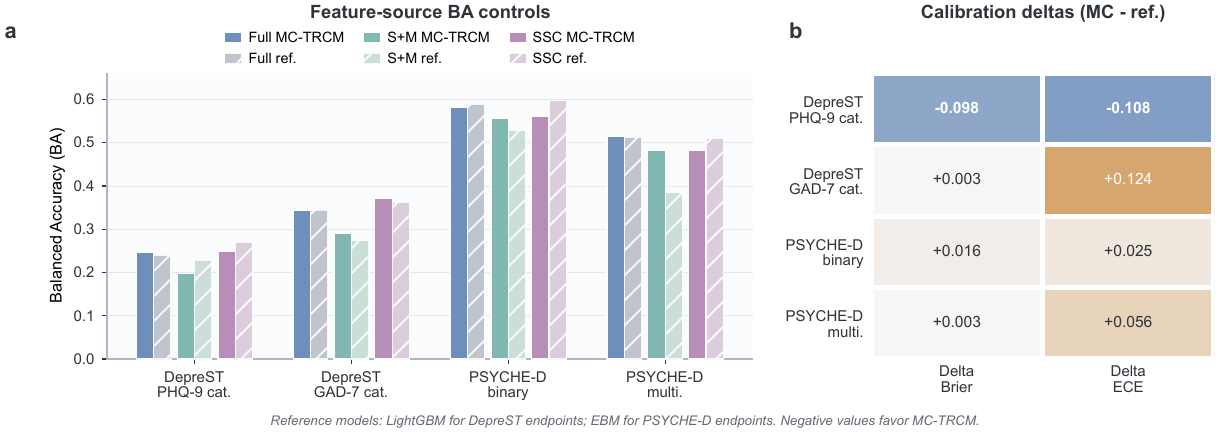}
\caption{Feature-source and calibration controls. \textbf{a} Balanced accuracy (BA) across feature-view subsets. Symptom/static/clinical context provides strong categorical signal, while sensor and missingness views contribute endpoint-dependent information. \textbf{b} Calibration deltas (MC minus reference); negative values favor \ModelName{}. Abbreviations: S+M (sensor + missingness), SSC (symptom/static/clinical context), and ref. (matched tabular reference).}
\label{fig:context_controls_nature_v2}
\end{figure}

Figure~\ref{fig:context_controls_nature_v2} places the main benchmark results in context. Panel~\textbf{a} shows that symptom, static, and clinical context carry substantial categorical signal: SSC-only results are competitive with, and for DepreST-CAT PHQ-9 category slightly above, the full \ModelName{} result. Sensor and missingness views alone are weaker on most endpoints, but the gap between S+M \ModelName{} and S+M references suggests that the proposed tokenization can extract more usable information from sparse observation patterns. Overall, the figure indicates that performance should not be attributed to passive sensing alone; released symptom and clinical context account for a meaningful share of the signal.

Panel~\textbf{b} shows endpoint-specific calibration behavior. \ModelName{} improves probability reliability on DepreST-CAT PHQ-9 category prediction, reducing Brier score and ECE by 0.098 and 0.108 relative to LightGBM. While some references show better calibration, we prioritize MAE and BA for their direct measure of clinical discriminative power—areas where \ModelName{} already demonstrates strong performance.

Recursive depth is selected on validation data from $K\in\{1,2,4,6,8\}$ using the method-defined validation score. Table~\ref{tab:recursive_depth_selection} shows that DepreST-CAT selects $K_d=8$, whereas PSYCHE-D peaks at $K_d=4$ and then declines. This non-monotonic behavior suggests that recursive refinement is useful but should be validation-locked rather than fixed a priori.

\begin{table}[!t]
\centering
\caption{Validation-selected recursive depth. Values are within-dataset validation selection scores used for recursive-depth tuning: severity endpoints contribute $1-\mathrm{MAE}/R_\tau$, and classification or ordinal endpoints contribute BA. Higher is better. Scores are not intended for comparison across datasets or as test metrics.}
\label{tab:recursive_depth_selection}
\scriptsize
\setlength{\tabcolsep}{3.2pt}
\renewcommand{\arraystretch}{1.05}
\begin{tabular}{@{}L{0.20\linewidth}C{0.09\linewidth}C{0.09\linewidth}C{0.09\linewidth}C{0.09\linewidth}C{0.09\linewidth}C{0.12\linewidth}@{}}
\toprule
Dataset & $K=1$ & $K=2$ & $K=4$ & $K=6$ & $K=8$ & Selected \\
\midrule
DepreST-CAT & 0.589 & 0.592 & 0.599 & 0.598 & \best{0.606} & 8 \\
PSYCHE-D & 0.548 & 0.552 & \best{0.563} & 0.555 & 0.553 & 4 \\
\bottomrule
\end{tabular}
\end{table}

\subsection{Ablation Studies}

Table~\ref{tab:ablation_compact} summarizes five-seed component ablations. The largest degradations come from removing FiLM and absence tokens, which increases severity MAE by 0.136 and 0.125, respectively, and also lowers category BA. This supports the main architectural claim that both context-conditioned fusion and explicit source absence are important for incomplete feature-view learning. Missingness projection and recursive refinement provide smaller but consistent gains, while PCGrad has only a minor effect in this protocol. Task conditioning mainly benefits severity estimation, with mixed average effects on categorical endpoints.

\begin{table}[!htbp]
\centering
\caption{Five-seed component ablations by metric family.}
\label{tab:ablation_compact}
\scriptsize
\setlength{\tabcolsep}{3.0pt}
\renewcommand{\arraystretch}{1.05}
\begin{tabular}{@{}L{0.34\linewidth}C{0.22\linewidth}C{0.22\linewidth}@{}}
\toprule
Variant & \makecell[c]{$\Delta$MAE\\(2 severity)} & \makecell[c]{$\Delta$BA\\(4 category)} \\
\midrule
Full MC-TRCM & 0.000 & 0.000 \\
No FiLM & +0.136 & -0.017 \\
No absence token & +0.125 & -0.012 \\
No missingness projection & +0.041 & -0.008 \\
No PCGrad & +0.008 & -0.003 \\
No task conditioning & +0.067 & +0.002 \\
No recursive refinement & +0.046 & -0.013 \\
\bottomrule
\end{tabular}
\par\smallskip
\begin{minipage}{0.96\linewidth}
\footnotesize\textit{Note.} Deltas are ablation minus Full; positive $\Delta$MAE indicates larger error, and negative $\Delta$BA indicates lower balanced accuracy.
\end{minipage}
\end{table}


\section{Limitations and Future Work}
This study identifies several deployment and generalization directions. First, calibration varies by endpoint. \ModelName{} improves calibration on DepreST-CAT PHQ-9 category prediction, while LightGBM has lower calibration error on GAD-7 category and EBM has lower Brier score and ECE on the two PSYCHE-D classification endpoints. These endpoint-level calibration patterns complement the discrimination results. Second, feature-source controls show that symptom and clinical context account for substantial signal in prepared public releases. This clarifies what passive sensing and missingness views add, rather than attributing performance to sensors alone.

Future work should extend \ModelName{} with uncertainty estimation, feature-source regularization, and domain adaptation across cohorts and release protocols. The current source-factorized tokenization can also be extended beyond pre-aggregated anchor summaries to process unaligned raw sensor time series when such streams are available.

\section{Conclusion}
\ModelName{} provides a structured observation-aware architecture for heterogeneous mobile and wearable mental-health feature views. Under validation-only selection, it achieves the lowest error on DepreST-CAT PHQ-9 and GAD-7 severity estimation and shows endpoint-specific classification behavior relative to strong tabular and neural references. Component ablations support the contributions of FiLM conditioning, missingness modeling, and recursive refinement, while feature-source accounting separates architecture effects from released symptom, static, and clinical context.

\begin{credits}
\subsubsection{\discintname}
The authors have no competing interests to declare that are relevant to the content of this article.
\end{credits}

\bibliographystyle{splncs04}
\bibliography{reference}
\end{document}